\documentclass[sigconf,nonacm]{acmart}

\usepackage{booktabs} 

\usepackage{tikz}     
\usepackage{bbm}   
\usepackage{url}      
\usepackage{amsmath}
\usepackage{algorithmic}
\usepackage{algorithm}
\usepackage{microtype}
\usepackage{subcaption}
\usepackage{xspace}

\usepackage{balance}
\usetikzlibrary{automata}        
\usepackage{lineno}
\usetikzlibrary{shapes,arrows, backgrounds}        
\usetikzlibrary{mindmap}
\usetikzlibrary{positioning}
\usetikzlibrary{decorations.pathreplacing}

\usetikzlibrary{shapes.geometric, arrows, positioning}

\usepackage{listings}

\definecolor{baseColor}{HTML}{03045e}
\definecolor{highlightColor}{HTML}{c9ada7} 
\definecolor{lightColor}{HTML}{ade8f4}

\tikzstyle{input} = [rectangle, rounded corners, minimum width=1cm, minimum height=1cm, text centered, draw=highlightColor, fill=highlightColor!30]
\tikzstyle{embedding} = [rectangle, rounded corners, minimum width=1cm, minimum height=1cm, text centered, draw=highlightColor, fill=baseColor!10, inner sep=5pt]
\tikzstyle{arrow} = [thick,->,>=stealth, draw=baseColor, dashed]

\tikzstyle{date} = [rectangle, fill=stateColor!10,text centered, rounded corners, text = stateColor, rotate=0]
\tikzstyle{eraBox} = [rectangle, fill= green!20,text centered,  text = stateColor, rotate=0]

\tikzstyle{latentVar} = [ellipse, draw, color = orange, fill= orange!0, text centered, text = orange]

\tikzstyle{obs} = [ellipse, draw, color = orange, fill= orange, text centered, text = white]

\newcommand{\cut}[1]{}

\newcommand{\spara}[1]{\smallskip\noindent{\bf{#1}}}

\newcommand{\eg}{{e.g.,}\xspace}

\newcommand{\ie}{{i.e.,}\xspace}

\theoremstyle{definition}
\newtheorem{example}{Example}

\usepackage{pifont}

\begin{document}
\title{Beyond the Hype: Embeddings vs. Prompting for Multiclass Classification Tasks}

\renewcommand{\shorttitle}{}

\fancyhead{}
%

\author{Marios Kokkodis}

\affiliation{%
  \institution{Thumbtack}
  \country{}
}
\email{mkokkodis@thumbtack.com}

\author{Richard Demsyn-Jones }
\affiliation{%
\institution{Thumbtack}
  \country{}
}
\email{demsynjones@thumbtack.com}

\author{Vijay Raghavan}
\affiliation{%
\institution{Thumbtack}
  \country{}
}
\email{vraghavan@thumbtack.com}

\begin{abstract}
Are traditional classification approaches irrelevant in this era of AI hype? We show that there are multiclass classification problems where predictive models holistically outperform LLM prompt-based frameworks. Given text and images from home-service project descriptions provided by Thumbtack customers, we build embeddings-based softmax models that predict the professional category (e.g., handyman, bathroom remodeling) associated with each problem description. We then compare against prompts that ask state-of-the-art LLM models to solve the same problem. We find that the embeddings approach outperforms the best LLM prompts in terms of accuracy, calibration, latency, and financial cost. In particular, the embeddings approach has 49.5\% higher accuracy than the prompting approach, and its superiority is consistent across text-only, image-only, and text-image problem descriptions. Furthermore, it yields well-calibrated probabilities, which we later use as confidence signals to provide contextualized user experience during deployment. On the contrary, prompting scores are overly uninformative. Finally, the embeddings approach is 14 and 81 times faster than prompting in processing images and text respectively, while under realistic deployment assumptions, it can be up to 10 times cheaper. Based on these results, we deployed a variation of the embeddings approach, and through A/B testing we observed performance consistent with our offline analysis. Our study shows that for multiclass classification problems that can leverage proprietary datasets, an embeddings-based approach may yield unequivocally better results. Hence, scientists, practitioners, engineers, and business leaders can use our study to go beyond the hype and consider appropriate predictive models for their classification use cases. 
\end{abstract}

%
%



\settopmatter{printacmref=false}

\maketitle


\section{Introduction}

AI hype is taking over the world: since the first version of OpenAI's ChatGPT there has been an increased focus and excitement on the advances of generative AI technology. AI companies' valuations have skyrocketed,\footnote{\url{https://www.cnbc.com/2024/12/27/how-five-top-ceos-described-the-ai-boom-in-2024.html}} accompanied by bold investments to push the boundaries towards Artificial General Intelligence. Big tech US companies are expected to invest a quarter of a trillion in AI innovation in 2025,\footnote{\url{https://www.forbes.com/sites/bethkindig/2024/11/14/ai-spending-to-exceed-a-quarter-trillion-next-year/}} more than ever before.

 In this environment, many tech companies are constantly striving to come up with AI use cases that would allow them to maintain relevant investment targets but also increase revenue, improve user experience, and, in general, provide better products and services. Naturally, this creates a working environment where leadership, product managers, software engineers, and designers propose AI-based solutions, primarily inspired by their individual experiences with prompt engineering. Paraphrasing the ``everyone is a programmer with AI'' quote,\footnote{\url{https://www.cnbc.com/2023/05/30/everyone-is-a-programmer-with-generative-ai-nvidia-ceo-.html}} everyone can propose machine learning (ML) solutions after chatting with an LLM.

Classification problems (\eg churn prediction~\citep{pamina2019effective}, fraud prediction~\citep{lokanan2022fraud}, and spam prediction~\citep{roy2020deep}) are some of the most common AI use-cases that can come up in business settings. Traditionally,  these problems are solved with classifiers such as logistic regression, gradient boosting, and multilayer perceptrons.  However, recently, with LLMs becoming more and more ubiquitous problem solvers, it is natural for many to think of and propose solutions that rely on prompt engineering~\citep{wang2023prompt, zhu2023prompt}.

For instance, and relevant to this work, assume that a company offers a keyword-based search function to its users, which they can use to search for products and services. Furthermore, assume that the company is interested in extending search capabilities by allowing flexible input (\eg natural text and/or images). Based on current trends and our discussion above, we can devise a prompt to solve this problem as follows:

\begin{example}
\label{example_problem}
\textit{Given a user's input search query and/or images, provide a list of products/services that the user is searching for. You need to choose products/services from this [{\tt predefined list of products and services}].}
\hfill $\blacksquare$
\end{example}

In the above, ``{\tt predefined list of products and services}'' refers to company-specific products and services that can be found through searching, and it is an input to the prompt. 

For traditional machine learning scientists, the above formulation of the problem as a prompting question might initially seem odd. Why would we use prompting for a clearly defined classification problem where there is company-specific data to train predictive models? For instance, we could first get numeric representations of text and images through a pre-trained model (e.g., OpenCLIP~\citep{ilharco_gabriel_2021_5143773}) and then build multiclass classification models on these representations to get a product or service prediction. Alternatively, we could also finetune a classifier from a pre-trained transformer-based model in a label-supervised manner~\citep{li2023labelsupervisedllamafinetuning}.

Yet, until this point, there is no clear evidence that a more traditional machine learning approach would be superior to prompting or vice versa. Prior results offer mixed guidance. A study on generic-knowledge public datasets showed that prompting might outperform fine-tuned models~\citep{gholamian2024llm}, while a different study on educational questions showed that embeddings-based approaches do better~\citep{al2024analysis}. Both studies, however, 
focus primarily on general knowledge problems and are limited to accuracy metrics. But what happens when company-specific definitions of products or services do not exactly match the general-knowledge text that LLMs were trained on? And beyond accuracy, companies typically use a more holistic evaluation framework to make a deployment decision that includes cost, latency, and calibration concerns. How do the two approaches compare within this more holistic evaluation framework? 

To fill this gap, we devise a study that compares the two approaches on a real search classification problem that we encountered at Thumbtack,\footnote{\url{https://www.thumbtack.com/}}, which is almost identical to Example~\ref{example_problem}. 
Thumbtack is a digital marketplace that connects homeowners with service professionals. The Thumbtack marketplace covers a wide variety of jobs, including renovations, repairs, maintenance, cleaning, and even various non-home projects. A typical customer scenario on Thumbtack starts with a customer typing some text in a search bar. Once the search query is submitted, Thumbtack's matching algorithm returns a list of relevant professionals that the customer can choose to contact and hire.

To improve user experience, Thumbtack is testing an expanded search capability that supports flexible input (i.e., natural text and/or images simultaneously). To do so, our team created an interface such that unstructured user input is ultimately translated into a task category from a predefined taxonomy of categories that Thumbtack uses to organize relevant professionals. At its core, identifying the most appropriate category is a multiclass classification problem, hence an ideal scenario to compare prompting with traditional classification approaches.

We start by experimenting with more than 10 different prompts and use OpenAI's\footnote{\url{https://openai.com/api/pricing/}} GPT-4o and GPT-4o mini to classify problem descriptions (customer input that includes natural text and images) into one of 95 frequent Thumbtack categories.  As an alternative, we use OpenCLIP~\citep{ilharco_gabriel_2021_5143773} to get numerical representations of text and images and then learn a softmax function to predict a probability distribution across the same 95 categories. 

We find that, compared to prompting, traditional ML approaches perform holistically better for this multi-class classification problem. In particular, our ML approach offers up to a 49.5\% accuracy improvement compared to the prompting approach, and its superiority is consistent across text-only, image-only, and text-image problem descriptions. Next, the ML approach yields relatively well-calibrated probabilities, which we later use as confidence signals to provide a contextualized user experience during deployment. In contrast, the prompt scores are overly uninformative. Furthermore, the ML approach is 14 and 81 times faster than prompting in processing images and text, respectively, while under realistic deployment assumptions, it can also be up to 10 times cheaper.

Based on the above results, we deployed a variation of the ML approach, and we ran an A/B test. The results of our experiment are on par with our offline analysis, showing similar predictive accuracies, calibrated probabilities, latency, and cost estimates.

Our work is the first to holistically compare traditional ML approaches with prompting for classification tasks on non-public data (i.e., data that LLMs were not trained on). Our results provide clear evidence of settings where traditional ML approaches are overly superior to prompting. As LLMs continue to advance and as they become ubiquitous tools in our everyday lives, it is important to showcase that the value that resides in proprietary data can often be well captured by more traditional ML approaches. As a result, our study can guide decision-making and can equip machine learning engineers and applied scientists with the necessary arguments to persuade cross-functional colleagues to go beyond the hype and deploy problem-appropriate approaches.

\section{Background}
\label{sec:bg}

A significant portion of relevant work has primarily focused on LLM applications and use cases such as API-calls~\citep{sheng2024measuring}, spelling correction~\citep{dutta2024enhancing},  commentary on live sport and music events~\citep{baughman2024large},  building agents for real-world tasks~\citep{fischer2024grillbot}, and detecting whether text was generated by AI~\citep{abburi2023generative}. 

In terms of prompting, some work focuses on label generation at scale~\citep{wan2024tnt}. For instance, ~\citeauthor{xu2024llms} showed that LLMs have clear limitations identifying the correct label when none of the correct answers is provided as a label option.  
Other prompting studies focus on providing frameworks for text classification. For example, \citeauthor{wang2023prompt} propose a prompt framework for improving zero-shot text classification,  while ~\citeauthor{zhu2023prompt} focus on optimizing prompting techniques to increase performance in short-text classification tasks. These studies, however, do not directly compare how training traditional ML models would perform for these tasks, which is the main focus of our work.

Closely related to our work, ~\citeauthor{al2024analysis} showed that a random forest built on embeddings outperformed prompting by 22\% at classifying educational questions, while ~\citeauthor{gholamian2024llm} found the opposite result, showing that few-shot prompting in fact outperforms supervised approaches. In contrast to our work, both studies use public datasets. Even if the exact data was not used to train the evaluated LLMs, those are datasets that align with publicly held conventions and definitions (e.g., the conventional definition of a camera, a ball, etc.). On the contrary, most company classification problems rely on proprietary data that often diverge significantly from publicly defined concepts (e.g., what does it mean for a customer to churn, what are the thresholds that fraud becomes a problem, what kind of problems can a handyman solve, etc.). For these types of problems, a supervised approach can clearly learn the patterns of proprietary data that might  not generalize to patterns found in public information used to train LLMs.

In addition, both studies are overly focused on accuracy. However, when deploying predictive classification models, accuracy is only one of the important dimensions--and arguably often not the most important one. Latency is of utmost importance since high latency can drive away customers~\citep{basalla2021latency}, while cost is also part of the decision process. Finally, calibration is another factor that companies care about in settings where prediction confidence affects decision-making ~\citep{Wang2023CalibrationID}. Our study covers all of these dimensions, hence filling the existing gap in related literature and providing a holistic guide for companies and practitioners on comparing prompting with traditional ML approaches.

\section{Problem formulation and methodology}
\label{sec:models}

At Thumbtack, one of the features that facilitates search and matching is the type of project that a customer is looking to complete. Over the years, Thumbtack has created a structured taxonomy that describes the types of tasks that customers typically search for on the platform. This taxonomy includes hundreds of project \emph{categories} that are used for efficient matching. 

In an effort to help customers express their needs in a more natural way, Thumbtack decided to invest in building a new search feature. The goal is to allow customers to use natural language and images to describe their problems. To achieve this, the first step requires an interface that would translate natural text and images into a relevant Thumbtack category that can be used to source professional services. Formally:

\vspace{0.2cm}
\spara{Problem definition:} \textit{Given a customer's natural language and/or image input,  return the most likely project category from a list of predefined Thumbtack categories.}
\\

To solve this classification problem we compare two approaches: prompting LLMs, and a more traditional embeddings-based predictive modeling approach. 

\subsection{Prompting approach}
\label{sec:prompt_approach}

Our zero-shot prompting approach relies on the assumption that state-of-the-art LLMs can be used to solve various classification problems~\citep{wang2023prompt,zhu2023prompt,al2024analysis,gholamian2024llm}. Zero-shot prompting, in which we task an LLM to solve a problem without any additional examples or training data, has several appealing properties. First, it is convenient for quick exploration, as you can set up a prompt within a few minutes. Second, it can create a baseline performance that can define the expectations of the hardness of the problem at hand. Finally, zero-shot prompting does not require any training data and hence can be particularly useful in cases where labeled data is not readily available or where general knowledge will likely outperform supervised approaches on predicting classes that rarely occur. 

Because of these properties, we wanted to quickly test the performance of the best available algorithms. At the time of our study, OpenAI's GPT-4o was the best available option~\citep{LLMLeaderboard}. However, given that it was more expensive and slower than its mini version, we decided to test whether GPT-4o mini was significantly worse than GPT-4o. As we show in Appendix~\ref{app:openai_comparison} and Figure~\ref{fig:openai_mini}, for our scenario, GPT-4o mini ended up outperforming GPT-4o. Hence, for the rest of the analysis we used GPT-4o mini.

One of our initial prompts is defined in the following Python format string:

\begin{lstlisting}
f'''Your goal is to identify the type of problem that a customer describes based on their description and any image(s) the customer shares. The type of the problem can be one of the following:  {ALL_CATS+['Other']} You will output a json object containing the 10 most likely types of problems given the user's input, in the following form:'''+'''{ 1st type of problem: float, 2nd type of problem: float,..., 10th type of problem: float }. The predicted scores should sum to 1.0, such that the most likely problem has a higher score than the second most likely problem, which should have a higher score than he third most likely problem, etc. Important: the most likely problems must be chosen from the {len(ALL_CATS)+1} problems listed in single quotes and separated by commas above.'''
\end{lstlisting}

For this prompt, we use OpenAI's {\tt v1/chat/completions} endpoint and we attempt to impose a JSON response structure. Furthermore, we instruct the LLM to make predictions within an accepted list of categories defined in variable {\tt ALL\_CATS}. (This version of the prompt had the highest performance within the {\tt v1/chat/completions} endpoint among 10+ similar alternatives.)

\begin{figure*}
\begin{center}
\begin{tikzpicture}[node distance=1cm]

\node (input) [input, text width=2cm, align=center]  {\underline{Input}: text description with Images};
\node (embedding) [embedding, right=of input, text width=2.5cm, align=center] {\underline{Embeddings model}: numerical representation of text and images};
\node (aggr) [embedding, right=of embedding, text width=2.5cm, align=center] {Embeddings aggregation (Algorithm~\ref{alg:agg})};
\node (decision) [embedding, right=of aggr, text width=2.7cm, align=center] {\underline{Classification model}: predict category};
\node (output) [input, right=of decision,  text width=2cm, align=center] {\underline{Output}: category to be used downstream (e.g., matching)};

\draw [arrow] (input) -- (embedding);
\draw [arrow] (embedding) -- (aggr);
\draw [arrow] (aggr) -- (decision);
\draw [arrow] (decision) -- (output);

\end{tikzpicture}
\caption{Predictive framework for category classification.}
\label{fig:framework}
\end{center}
\end{figure*}

The above prompt uses natural language to guide the LLM to provide a response from a predefined list of categories. This approach does not enforce structure and as such does not guarantee that the prompt will return a valid response. To fix this issue, we use a Pydantic-based prompt along with OpenAI's structured output functionality within the {\tt beta.chat.completions.parse} function. In particular, we first define the following enum class of available categories:

\begin{lstlisting}
class ServiceCategory(Enum):
    Appliance_Installation = 'Appliance Installation'
    Balloon_Decorations = 'Balloon Decorations'
    # Other categories here
\end{lstlisting}

\noindent Then, we define the following response class:

\begin{lstlisting}
class ServiceCategoryPrediction(BaseModel):
    predictions: list[ServiceCategory]
    scores: list[int]
\end{lstlisting}

Finally, we force the LLM  to provide a {\tt ServiceCategoryPrediction} response, hence ensuring that the category predictions will be within the defined category members of the {\tt ServiceCategory} class. 
The best-performing prompt, which is the one we use for the rest of the paper, is the following:

\begin{lstlisting}
'''Your goal is to identify the type of problem that a customer faces based on the customer's text description and any image(s) the customer shares. The type of the problem must be a member of the ServiceCategory enumeration class. You will output the 10 most likely ServiceCategory enumeration members given the user's input along with their likelihood scores. The likelihood scores should be in a scale from 1 to 10, and they should represent your confidence such that the most likely ServiceCategory enumeration member has a higher integer score than the second most likely, which should have a higher score than the third most likely, etc. 
'''
\end{lstlisting}

\subsection{Embeddings-based approach}

A more traditional approach to this problem is to define text and/or image representations and then try to learn a model that predicts the category based on these representations. 
An efficient transformer to achive this is 
OpenClip~\citep{radford2021learning, ilharco_gabriel_2021_5143773}. OpenClip models are trained by predicting image captions they can learn to represent both text and images in the same contextual space.\footnote{https://huggingface.co/docs/hub/en/open\_clip} 

For our use case, we design the predictive framework shown in Figure~\ref{fig:framework}. In particular, we start by passing the input through the OpenClip model. (For this implementation, we used OpenCLIP VIT-g-14, but the results were almost identical for lighter versions, such as the VIT-L-14.) The output of this step is a list of vectors that represent the input text and images. Since the length of this list depends on the number of images, we choose to aggregate (pool~\citep{gholamalinezhad2020pooling}) them into a single N-dimensional vector as shown in 
Algorithm~\ref{alg:agg}. It is possible that weighting image embeddings differently than text embeddings could lead to better models (Section~\ref{sec:discussion}), however, for the purposes of this study, we wanted to build the simplest vanilla embeddings approach possible. 

\small
\begin{algorithm}
\caption{Aggregation of image and text embeddings }
\label{alg:agg}

\begin{algorithmic}[1]
\STATE \textbf{Function} \textsc{image\_repr}(images)
\STATE \hspace{1em} \textbf{Input:} A list of images
\STATE \hspace{1em} \textbf{Initialize} image\_emb $\gets [0] * $ OpenClip dimension
\STATE \hspace{1em} \textbf{For each} image \textbf{in} images \textbf{do}
\STATE \hspace{1em} \hspace{1em} image\_emb $\gets$ result $+$ OpenClip(image)
\STATE \hspace{1em} \textbf{End for}
\STATE \hspace{1em} \textbf{Return} image\_emb / $\text{len(images)}$

\STATE \textbf{Function} \textsc{text\_repr}(text)
\STATE \hspace{1em} \textbf{Input:} A text description
\STATE \hspace{1em} \textbf{Return} \texttt{OpenClip(text)}

\STATE \textbf{Procedure} \textsc{AGG\_Representations}(text, images)
\IF{text is null}
    \STATE \textbf{Return} \textsc{image\_repr}(images)
\ELSIF{images is null}
    \STATE \textbf{Return} \textsc{text\_repr}(text)
\ELSE
    \STATE \textbf{Return} $\mathbf{X}$ = $\frac{\textsc{image\_repr}(images) + \textsc{text\_repr}(text)}{2}$
\ENDIF

\end{algorithmic}
\end{algorithm}
\normalsize

We construct a corpus where we treat the category of the professional chosen by the user as the ground truth label for each observation. We predict this label with a multiclass classification model using the output of the aggregation algorithm as its input. An important assumption here is that a professional who was contacted by a user met the user's problem requirements or was in the most suitable category to do so. Assuming a set of $C$ categories, we estimate a logistic regression model of the following form:

\begin{equation}
\label{eq:logit}
    \Pr(Y_i = c | \mathbf{X}_i; \boldsymbol \beta) = \frac{\exp{\boldsymbol \beta_c \mathbf{X}_i}}{\sum_{j \in C} \exp{\boldsymbol\beta_j \mathbf{X}_i}} \;,
\end{equation}
where $\mathbf{X}_i$ is the embeddings aggregated vector (line 17 in Algorithm~\ref{alg:agg}) of observation $i$, $c$ is the ground truth category, and  $\boldsymbol \beta$ is the vector of coefficients that we need to estimate. 

For category prediction, we return the most likely category as follows:
\begin{equation}
    \arg\max_{j \in C} \Pr(Y_i = j | \mathbf{X}_i; \boldsymbol \beta)
\end{equation}
For evaluating how calibrated our approach is and comparing it with the scores that prompting returns we use the actual output probabilities.

\section{Experimental Setup and Results}
\label{sec:exp}
Next, we discuss the dataset and the setup we used to compare the two approaches.

\subsection{Data}
\label{sec:data}

For our analysis, we compiled a randomly selected set of 45,122 problem descriptions provided by Thumbtack customers. We split the dataset over time, with 37,634 observations forming the train set and 7,488 subsequent observations forming the test set. All problem descriptions are from Fall 2024, with observations in the training set occurring before September 2024 while all problem descriptions in the test set occurred afterwards. We use the training set to learn the parameters of Equation~\ref{eq:logit}. We use the test set to evaluate the performance of both approaches. 

\begin{figure}
\begin{center}
\includegraphics[width=0.5\textwidth]{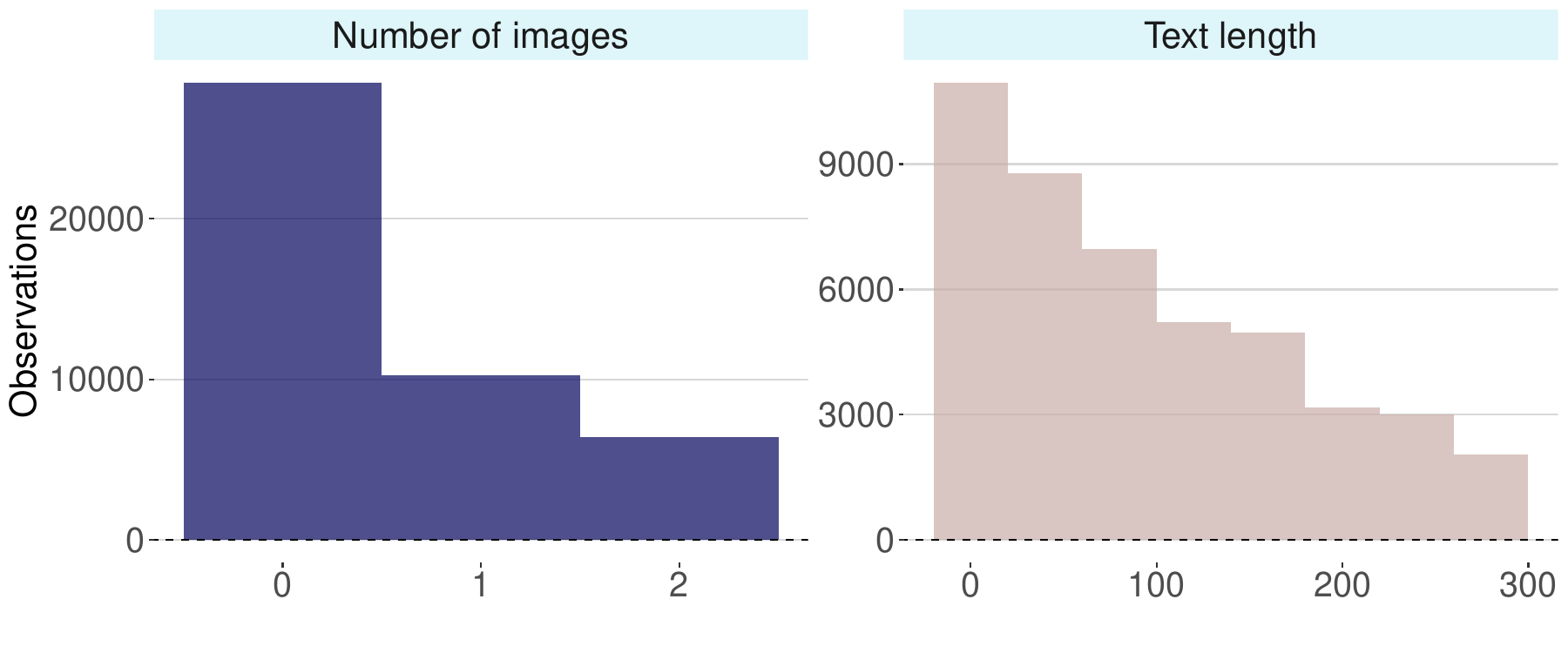}
\vspace{-0.3cm}
\caption{The distributions of the number of images and text length in our dataset.}
\label{fig:distros}
\end{center}
\end{figure}

The problem descriptions can be text-only, or image-only, or they can include both text description and image(s). Users provide these problem descriptions to their chosen professionals through an input text field, and those professionals can request more information or quote prices for the work (e.g., how much it would cost to mount a TV, fix a toilet, or remodel a bathroom). Figure~\ref{fig:distros} provides an overview of the input distributions for the number of images and the length of the text.  Table~\ref{tab:dataset} provides the breakdown of our dataset across these different types of inputs. 
As we show in the figure,  we focus on problem descriptions that had up to 2 images and less than 300 characters of text. 

Thumbtack offers several hundreds of project categories, from handyman and house cleaning tasks to piano lessons and logo design. For the purposes of this study, we focus on the 95 most frequent categories. However, note that we also experimented with fewer and larger sets of categories, and while qualitatively the results were similar, we observed a significant decrease in prompt performance compared to the embeddings approach as the number of categories increased.

\begin{figure}
\begin{center}
\includegraphics[width=0.43\textwidth]{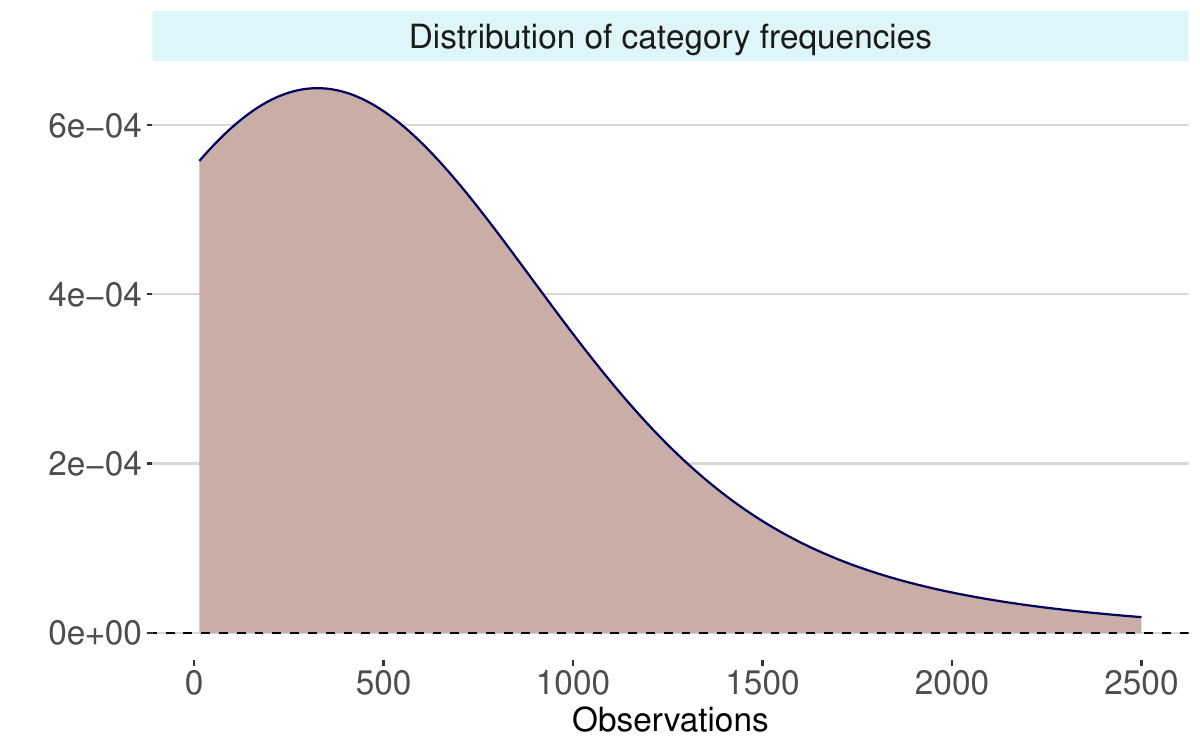}
\vspace{-0.3cm}
\caption{The frequency distribution of the 95 categories we consider in this study.}
\label{fig:label_distros}
\end{center}
\end{figure}

Figure~\ref{fig:label_distros} shows the distribution of the 95 categories (\ie the classes of our target variable--capped at 2500 observations. The majority of these categories are observed roughly 200 times in our dataset, with a few frequent ones generating a tail in the input distribution.

\subsection{Evaluation metrics}
We compare the two approaches in terms of accuracy, predicted probabilities,  latency, and cost. 

\spara{Accuracy}: We use both accuracy@k and a variation of accuracy@k to benchmark the accuracy of the two approaches. For accuracy@k we use the following formula:

\begin{equation}
    \text{Accuracy@}k = \frac{1}{|T|} \sum_{i\in T} \sum_{j=1}^k 1_{Y_i=\hat{Y}_{ij}} \;, \; 
    \label{eq:acc}
\end{equation}
where $1_{Y_i=\hat{Y}_{ij}}$ is an indicator function that is 1 when the $j-th$ ranked prediction for observation $i$, $\hat{Y}_{ij}$, is equal to the correct category of $i$ and 0 otherwise, and $k$ is the number of top-ranked predictions we consider.

Accuracy@k is a good metric, however it misses correlations between closely related categories where relevance may not be binary. For instance, assume a scenario where $Y_i=\textit{TV mounting}$  and $\hat{Y}_{ij}=\textit{handyman}$. A handyman might be able to perform a TV mounting job, so it would be misleading for this prediction to have the same error as if $\hat{Y}_{ij}=\textit{swimming lessons}$. 
Our data allows us to estimate the {\bf relevance} of a category $p$ to another category $q$ by sampling across our professionals and calculating how often they offer services in both categories $p,q$.

Let $P$ and $Q$ be the set of professionals who offer services in categories $p$ and $q$, respectively. We define the relevance of $q$ to $p$ as:

\begin{equation}
\text{Rel}(p, q) =
\frac{|P \cap Q|}{|P|}
\end{equation}

We calculate relevance between categories using a representative sample of Thumbtack professionals separate from our focal dataset. Then, we  use this relevance score to estimate a more {\bf relative} accuracy@k metric as follows:
\begin{equation}
    \text{Relative Accuracy@}k = \frac{1}{|T|} \sum_{i\in T} \min\Big(1,\sum_{j=1}^k Rel (Y_i, \hat{Y}_{ij})\Big) \;, 
    \label{eq:rel}
\end{equation}
where $T$ is the test set and $Rel(Y_i, \hat{Y}_{ij})$ captures how relevant the predicted category $\hat{Y}_{ij}$ is to the actual category $Y_i$.

\begin{table}
\small
\begin{center}
\begin{tabular} 
 { l l l  } 
 \toprule 
 & 
 \bf Train & \bf Test  \
 \\
 \midrule
\bf Image-only   & 2,870 &  451
\\ 
\bf Image and text   & 10,935  & 2,407
\\ 
\bf Text-only  & 23,829 &  4,630
\\ 
\midrule
\bf Total  & 37,634 &  7,488
\\ 
\bottomrule 
 \end{tabular} 
 \vspace{0.2cm}
 \caption{Train and test datasets breakdown across types of inputs. Train and test sets are split over time, with the test set occurring right after the end of the train set.} 
\label{tab:dataset}
\end{center}
\vspace{-0.7cm}
\end{table}

\begin{example}
\textit{Assume that we have the following top prediction: $\hat{Y}=\textit{Appliance Installation}$ while the actual true category is \textit{Furniture Assembly}. Looking into our data we observe that 80\% of the professionals who offer \textit{Furniture Assembly} services also offer \textit{Appliance Installation}. Hence, in this example:}

\begin{eqnarray*}
\text{Accuracy@1} &=& 0 \\ 
\text{Relative Accuracy@1} &=& 0.8
\end{eqnarray*}
\hfill $\blacksquare$
\end{example}

It should be clear by the definition of Equation~\ref{eq:rel} that this quantity is asymmetric, where typically $Rel(p,q) \neq Rel(q,p)$. This is by construction since we are only interested in the likelihood that a professional can offer a predicted category $\hat{Y}$ when the actual category is $Y$ and not vice versa. In addition, note that relative accuracy can {\em never be greater than 1}. In cases where the total relevancy between the top $k$ predicted categories  and the ground truth category exceeds 1, relative accuracy will be 1 through the application of the $\min$ function in Equation~\ref{eq:rel}.

\begin{figure}
\begin{center}
\includegraphics[width=0.48\textwidth]{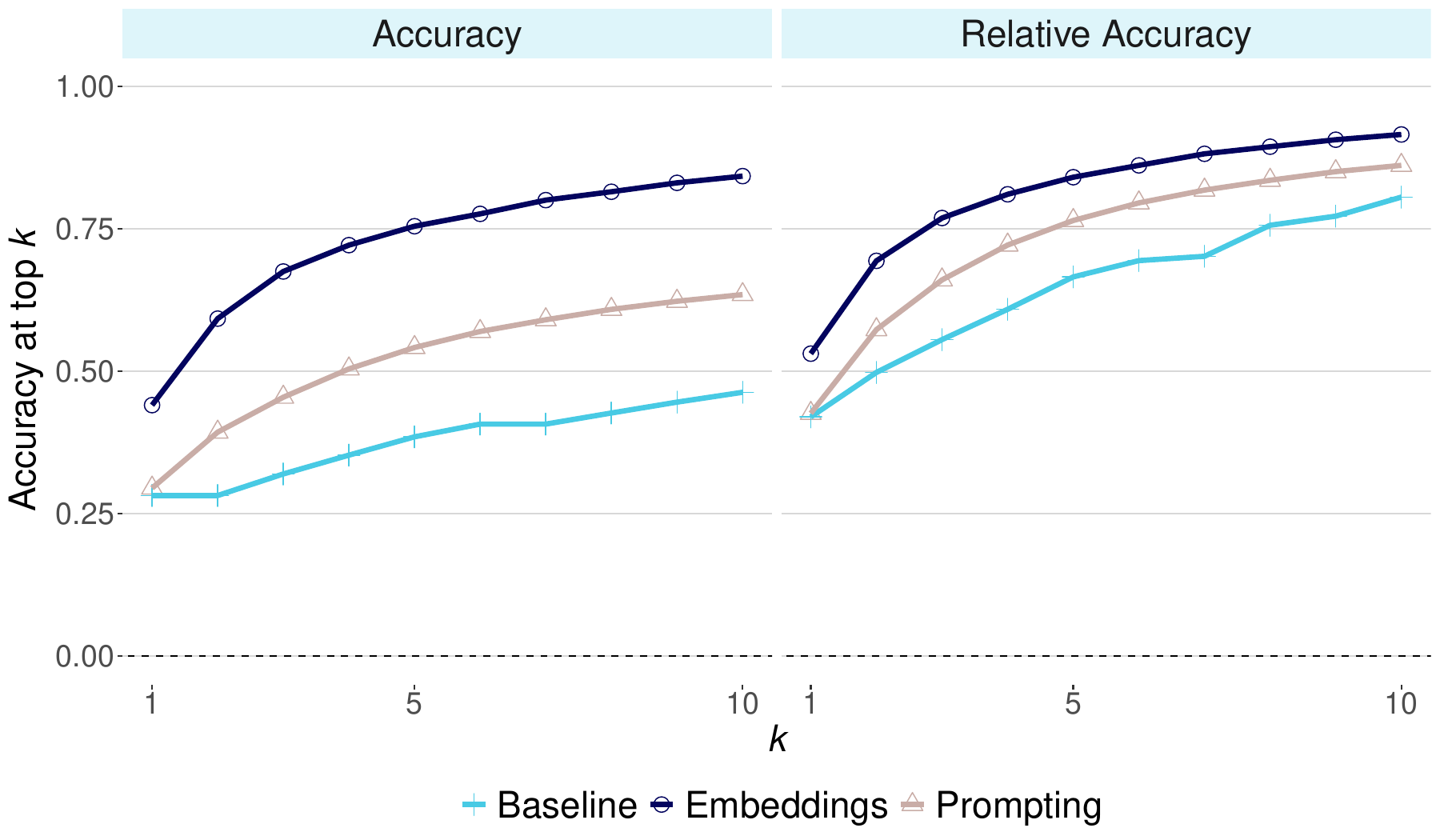}
\vspace{-0.3cm}
\caption{The embeddings approach outperforms prompting in terms of accuracy @ (Equation~\ref{eq:acc}) and relative accuracy @ (Equation~\ref{eq:rel}).}
\label{fig:acc_topn}
\end{center}
\end{figure}



\spara{Calibration}:  Understanding the confidence of our predictions is paramount, as we can choose to ask users for more information or we can provide alternative categories for them to choose from. 
Hence, evaluating how calibrated our probabilities are is particularly important. 

Since we have a multiclass classification problem, we calculate the Brier Score loss\footnote{\url{https://en.wikipedia.org/wiki/Brier_score}} as follows:

\begin{equation}
 \text{Brier Score loss}={\frac {1}{|T|}}\sum \limits _{t=1}^{T}\sum \limits _{c=1}^{C}(f_{tc}-o_{tc})^{2}\,\!
 \label{eq:brier}
\end{equation}
where $C=95$ classes in our problem, $f_{tc}$ is the predicted probability for instance $i$ to be class $c$, and $o_{tc}$ is 1 when the actual category for instance $t$ is $c$ otherwise is zero. 

In the scenario where we primarily care about the most likely prediction (\ie if we want to match customers with professionals from the most likely category) we would like to know how well-calibrated the probabilities of our top predictions are. One way to evaluate this is to draw calibration graphs and visually compare them with the ideal $y=x$ calibration line. 

\spara{Latency}: The third dimension we consider is latency. To estimate the response speeds of each approach, we first create a randomly selected sample of 100 images and a separate sample of 100 texts. For the prompting approach, we calculate the average time between a request and a response for each one of the images or texts in each sample. Similarly, for the embeddings approach, we calculate the average time that it takes to calculate a representation and make a prediction.

\begin{figure}
\begin{center}
\includegraphics[width=0.48\textwidth]{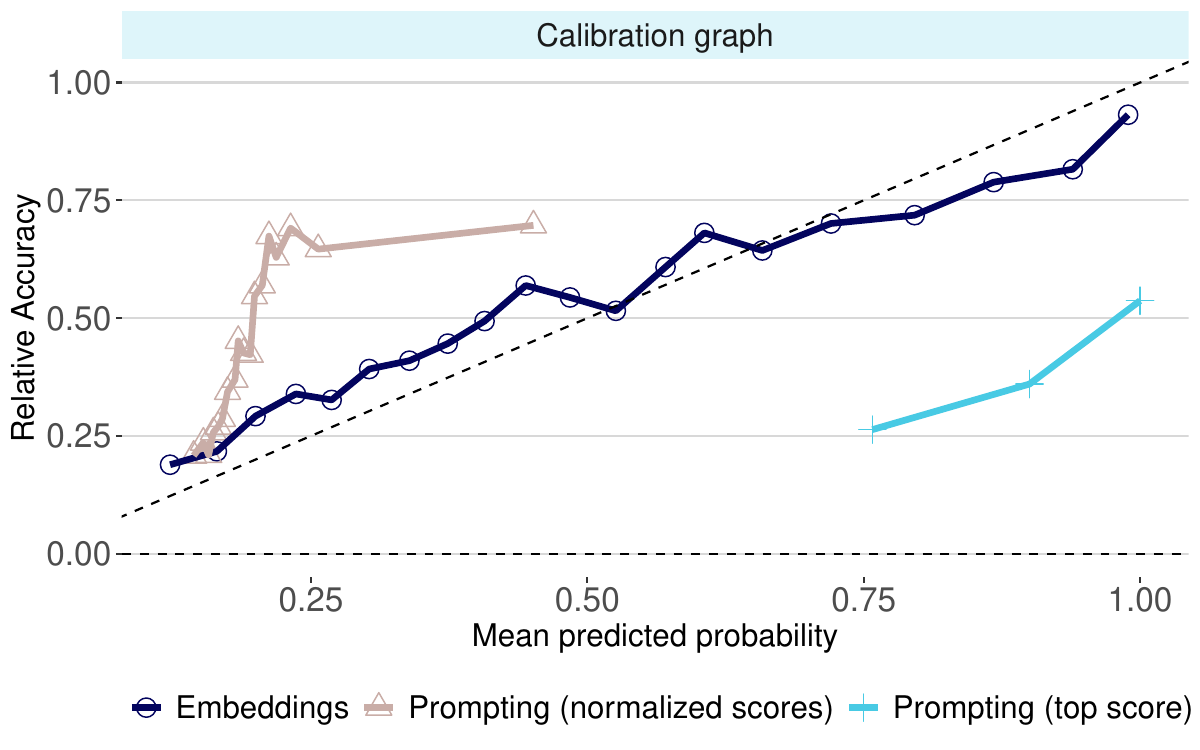}
\vspace{-0.3cm}
\caption{The embeddings approach provides significantly more calibrated probabilities compared to prompting.}
\label{fig:calib}
\end{center}
\end{figure}

\spara{Cost}: Finally, to calculate the cost of each approach, we rely on the official OpenAI\footnote{https://openai.com/api/pricing/} and AWS x2gLarge\footnote{\url{https://instances.vantage.sh/aws/ec2/g6.xlarge}} pricing,
and we assume equal traffic volume and a blue/green deployment\footnote{https://docs.aws.amazon.com/whitepapers/latest/overview-deployment-options/bluegreen-deployments.html} for the embedding-based approach. 

\subsection{Results}
\label{sec:results}
Below we report the performance of the two approaches. 

\spara{Accuracy:} We estimate the accuracy @ and relative accuracy @ of each approach according to Equations~\ref{eq:acc} and~\ref{eq:rel}. Besides the focal approaches, we also estimate the accuracy of a baseline approach, i.e., a majority classifier that always predicts the top $k$ most frequent categories. Figure~\ref{fig:acc_topn} shows the results for $k \in \{1, 2, ..., 10\}$. First, we observe a clear and consistent superiority of the embeddings approach compared to the prompting and the baseline approaches. In terms of accuracy, at $k=1$, the embeddings approach scores 44.1\%, a 49.5\% improvement over the prompting approach, which scores 29.5\%,  barely beating the baseline (28.2\%). Similarly, in terms of relative accuracy, the embeddings approach scores 53.1\%, 24.6\% better than the prompting approach (42.6\%), which again performs on par with the baseline (42\%). 
Furthermore, across both accuracy metrics, the embedding approach maintains its superiority over the prompting approach and the baseline as $k$ increases from 1 to 10.

If we break down the performance of the two approaches across the different types of inputs shown in Table~\ref{tab:dataset}, we observe that the embeddings approach outperforms the prompting approach across all three subcategories and across all $k$ (Figure~\ref{fig:text_image}, Appendix~\ref{app:inputs}). Focusing only on $k=1$ and for text and images, the embeddings approach has an accuracy of 56.2\%, 37\% better than prompting's 41\%. Similarly, for image-only problem descriptions, embeddings score 41.2\% at $k=1$, 24.5\% better than prompting's 33.1\%. For text-only problem descriptions, embeddings score 37.3\%, 64\% better than prompting's 22.8\%. The results are similar across all subgroups in terms of relative accuracy.

\spara{Calibration}: To examine how meaningful the predicted probabilities of the two approaches are, we first estimate the Brier Score loss (Equation~\ref{eq:brier}).  Our prompting approach does not return probability scores; instead, it rates its confidence from 1 to 10 (see second prompt in Section~\ref{sec:prompt_approach}). We purposely switched from probabilities to integers because prompting's performance with decimal numbers was significantly worse. This switch is shown in our progression from the first to the second prompt of Section~\ref{sec:prompt_approach}. One way to create probability distributions where all scores sum to 1 is to divide all scores by their sum across each observation, under the assumption that relative differences in confidence scores should equally represent relative differences in probability.

By performing the above normalization, we find that the embeddings approach has a multiclass Brier Score loss equal to 0.78, significantly lower than the Brier Score loss of prompting (0.93).

Figure~\ref{fig:calib} shows how the relative accuracy @ $k=1$ (Equation~\ref{eq:rel}) of the most likely prediction changes as the mean predicted probability of different quantiles increases (x-axis). This is a calibration graph of the most likely probability of our set of multiclass probabilities, except for the y-axis showing relative accuracy @ $k=1$ instead of accuracy @ $k=1$.\footnote{Recall that relative accuracy is by construction bounded in [0,1].} 

Despite these two differences, the interpretation of the graph is equivalent to a calibration graph: a perfectly calibrated model will have a straight line with a  45 degree slope. For the embeddings approach, we observe a relatively well-calibrated graph, with points above and below the perfect dotted line. 
For prompting, using the same normalization we used for estimating the Brier Score loss, we observe a concentration of the predicted probabilities in the top left corner of the graph. For this evaluation, since we are only interested in the most likely prediction, we also estimate the probabilities of the top prediction (predicted top score over 10). As we show in the figure,  all top predicted scores concentrate in the bottom right corner of the graph.

\begin{figure}
\begin{center}
\includegraphics[width=0.48\textwidth]{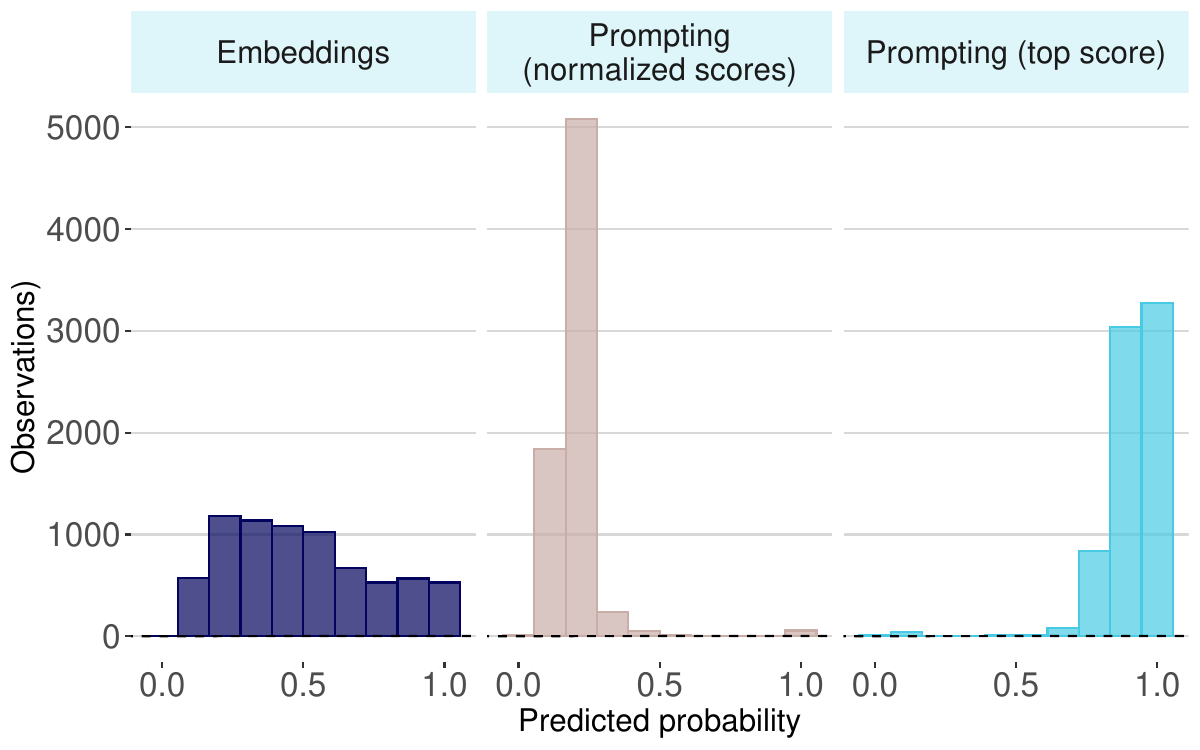}
\vspace{-0.3cm}
\caption{While the embeddings approach predicts probabilities between 0 and 1 in an almost uniform fashion, prompting returns concentrated scores that render its predictions mostly uninformative.}
\label{fig:predicted_probs}
\end{center}
\end{figure}

Figure~\ref{fig:calib} suggests that in terms of prompting predicted scores, there is not enough variation. To evaluate, in Figure~\ref{fig:predicted_probs} we show the distributions of predicted probabilities of the embeddings and the two prompting approaches. The figure clearly shows that prompting offers little variation in its predictions despite the immense range of category types supported on Thumbtack's platform. This suggests that we cannot use prompting predicted scores as a signal of confidence. In contrast, the embeddings approach provides a close-to-uniform predicted probability distribution, allowing us to interpret these scores as confidence and to tune the behavior of our deployed search feature appropriately.

\spara{Latency}: The average time that it takes for an image to be processed by our embeddings approach on an NVIDIA L4 GPU  is 0.3 seconds (95\% Confidence Interval (CI): [0.28, 0.32]), while, on average, it takes 14 times longer for the average OpenAI  response (95\% CI: [3.77, 4.5], with an average of 4.14 seconds). 
Similarly, for text descriptions, it takes 0.0153 seconds (95\% CI: [0.0148, 0.0158]) for the embeddings approach while it takes 1.22 (95\% CI: [1.10, 1.35]) seconds for OpenAI to respond.  For any interactive user product, expecting users to wait 4 seconds to receive a response can have a negative revenue impact~\cite{basalla2021latency}. For many of the original problem descriptions with images we parsed, the waiting time was significantly larger than that, reaching up to 11 seconds to receive a response. Response time significantly decreases if we apply image downsizing, but still it remains on average above 2 seconds for images while accuracy also decreases.


\spara{Cost}: 
We can contrast an internal deployment utilized at full capacity against the equivalent costs of OpenAI API calls.
A single node of an L4 GPU can process approximately 3 images per second or 65 text queries. For calculation purposes, let's assume that a single node can process 2 text+image requests per second. Hence, in a day, a single node can process 86,400 * 2 = 172,800 queries with images and texts. Over a year, a single node can process 63M requests. The cost given a blue/green deployment (i.e., we would need 2 nodes) will be 2 * \$4,589.54 = \$9,179 per year.\footnote{\url{https://instances.vantage.sh/aws/ec2/g6.xlarge}} 

On the other hand, if we assume that we will use the cheapest GPT-4o mini model, and that we will resize images to 250px * 250px (roughly) then the annual cost for parsing 63M requests will be 63M * \$0.001275 = \$80,325. In addition, our prompt is roughly 1,270 tokens (including expected user input of ~10 tokens). OpenAI charges \$0.150 per 1M tokens. This would cost
(63M * 1,270 * \$0.150)/1M = \$12,000. Hence, the total OpenAI cost for GPT-4o mini will be \$92,325, which is 10x more expensive. 

An important caveat is that this cost comparison assumes full capacity utilization of the GPU, and that GPUs are provisioned in discrete quantities. For volumes that would imply under-utilization of one or more GPUs, the cost comparison can be less dramatic;  regardless, except for extremely low volumes of traffic, provisioned resources will be cost-effective compared to OpenAI API calls.


 \spara{Summary:} Overall, the above analysis outlines the superiority of the embeddings-based approach. In terms of accuracy, embeddings perform by up to 49.5\% better than prompting, consistently across text-only, image-only, and text-and-image-only problem descriptions. In terms of calibration, embeddings provide informative scores that can be used as confidence signals, while prompting provides mostly uninformative scores. In terms of latency, embeddings can provide significantly faster responses, while in terms of cost, embeddings are significantly cheaper.

\section{Discussion}
\label{sec:discussion}

In this work, we showed that there are multiclass classification problems where predictive models holistically outperform LLM prompt-based frameworks. By comparing an embeddings-based approach with prompting, we found that the former significantly outperforms the latter across accuracy, calibration, latency, and financial cost. Based on these results, we deployed a variation of the ML approach, and through A/B testing we observed performance consistent with our offline analysis.

\spara{Contributions}: Our study shows that for multiclass classification problems that can leverage proprietary datasets, an embedding-based approach may unequivocally yield better results. While prior studies using public datasets showed conflicting results (Section~\ref{sec:bg}), our study provides for the first time clear evidence that, especially with proprietary data, there is little to argue for a prompting-based approach: a supervised approach can better learn the patterns of company-specific data that do not necessarily generalize to outside conventions. Even further,  our work illustrates how poorly prompting approaches perform in terms of score prediction, which is paramount for deploying different user experiences based on confidence. 
Finally, our study is the first to discuss cost and latency characteristics of the two approaches, both of which are particularly important when deploying applications.

\spara{Limitations and future directions}:
Our study is not without limitations. First, latency and cost could be lower if we deployed any of the open-source models (e.g., DeepSeek~\citep{guo2025deepseek}, LLama~\citep{touvron2023llama}) internally. However, this approach has its own shortcomings. First, deploying internally removes much of the speedy experimentation advantage that prompting has (since internal deployment requires similar effort as deploying a pre-trained embeddings transformer). Second, one of the benefits of calling OpenAI is that we can get the necessary structured output (Section \ref{sec:prompt_approach}). In an internal deployment, we would have to spend time developing a filtering layer or add dependencies such as LangChain\footnote{\url{https://www.langchain.com/}} 
to enforce the required output.  

One approach that we did not evaluate is whether RAG-based prompting would have yielded better accuracy results. It is rational to assume that feeding proprietary data in an LLM could improve performance. Yet, a RAG-based approach would further increase the latency and cost difference between the two approaches, while it is unlikely to improve the probability score predictions. 

Another variation that we did not try is few-shot prompting, where we could provide examples that could teach the LLM to make better predictions. Similar to a RAG-based approach, few-shot prompting could improve LLM accuracy by using proprietary data (albeit requiring a long context model to include substantial input text in cases similar to ours where the number of predicted classes is in the hundreds). However, few-shot prompting would likely increase both latency and cost: some long prompts we tested took up to 20 seconds to complete, rendering them unrealistic for real-time deployment. Even further, and similar to the aforementioned alternatives, few-shot prompting would likely not resolve prompting's calibration limitations discussed in this work. 

Despite these shortcomings, we hope that our study will incentivize other researchers to explore few-shot and RAG-based approaches and holistically compare them with supervised machine learning frameworks trained on proprietary data. 

It is important to highlight that our study does not argue that prompting is generally bad. Instead, we focus on a particular application of prompting (multiclass classification) that has recently gained traction.  Prompting is great and we use it constantly at Thumbtack for summarization and other types of unstructured natural language tasks where outside information is sufficient. However, when it comes to classification problems with labelled data, we believe that prompting is not yet comparable to traditional approaches.

Finally, in this work we used a vanilla logistic regression application that relied only on embeddings. Alternative approaches could use additional features (e.g., length of text), different types of weighting images and texts, image and text-specific models in a mixture of expert frameworks, and models such as gradient boosting or random forests, or even fine-tuned transformers.

Overall, besides these limitations, we are confident that our study can inform scientists, practitioners, engineers, and business leaders to go beyond the hype and consider appropriate predictive models for their classification use cases.

\bibliographystyle{ACM-Reference-Format}
\bibliography{em}


\begin{thebibliography}{23}


\ifx \showCODEN    \undefined \def \showCODEN     #1{\unskip}     \fi
\ifx \showDOI      \undefined \def \showDOI       #1{#1}\fi
\ifx \showISBNx    \undefined \def \showISBNx     #1{\unskip}     \fi
\ifx \showISBNxiii \undefined \def \showISBNxiii  #1{\unskip}     \fi
\ifx \showISSN     \undefined \def \showISSN      #1{\unskip}     \fi
\ifx \showLCCN     \undefined \def \showLCCN      #1{\unskip}     \fi
\ifx \shownote     \undefined \def \shownote      #1{#1}          \fi
\ifx \showarticletitle \undefined \def \showarticletitle #1{#1}   \fi
\ifx \showURL      \undefined \def \showURL       {\relax}        \fi
\providecommand\bibfield[2]{#2}
\providecommand\bibinfo[2]{#2}
\providecommand\natexlab[1]{#1}
\providecommand\showeprint[2][]{arXiv:#2}

\bibitem[Abburi et~al\mbox{.}(2023)]%
        {abburi2023generative}
\bibfield{author}{\bibinfo{person}{Harika Abburi}, \bibinfo{person}{Michael Suesserman}, \bibinfo{person}{Nirmala Pudota}, \bibinfo{person}{Balaji Veeramani}, \bibinfo{person}{Edward Bowen}, {and} \bibinfo{person}{Sanmitra Bhattacharya}.} \bibinfo{year}{2023}\natexlab{}.
\newblock \showarticletitle{Generative ai text classification using ensemble llm approaches}.
\newblock \bibinfo{journal}{\emph{arXiv preprint arXiv:2309.07755}} (\bibinfo{year}{2023}).
\newblock


\bibitem[Al~Faraby et~al\mbox{.}(2024)]%
        {al2024analysis}
\bibfield{author}{\bibinfo{person}{Said Al~Faraby}, \bibinfo{person}{Ade Romadhony}, {et~al\mbox{.}}} \bibinfo{year}{2024}\natexlab{}.
\newblock \showarticletitle{Analysis of llms for educational question classification and generation}.
\newblock \bibinfo{journal}{\emph{Computers and Education: Artificial Intelligence}}  \bibinfo{volume}{7} (\bibinfo{year}{2024}), \bibinfo{pages}{100298}.
\newblock


\bibitem[Analysis({[n.\,d.]})]%
        {LLMLeaderboard}
\bibfield{author}{\bibinfo{person}{Artificial Analysis}.} \bibinfo{year}{[n.\,d.]}\natexlab{}.
\newblock \bibinfo{title}{LLM Leaderboard}.
\newblock
\newblock
\urldef\tempurl%
\url{https://artificialanalysis.ai/leaderboards/models}
\showURL{%
\tempurl}


\bibitem[Basalla et~al\mbox{.}(2021)]%
        {basalla2021latency}
\bibfield{author}{\bibinfo{person}{Marcus Basalla}, \bibinfo{person}{Johannes Schneider}, \bibinfo{person}{Martin Luksik}, \bibinfo{person}{Roope Jaakonm{\"a}ki}, {and} \bibinfo{person}{Jan Vom~Brocke}.} \bibinfo{year}{2021}\natexlab{}.
\newblock \showarticletitle{On latency of e-commerce platforms}.
\newblock \bibinfo{journal}{\emph{Journal of Organizational Computing and Electronic Commerce}} \bibinfo{volume}{31}, \bibinfo{number}{1} (\bibinfo{year}{2021}), \bibinfo{pages}{1--17}.
\newblock


\bibitem[Baughman et~al\mbox{.}(2024)]%
        {baughman2024large}
\bibfield{author}{\bibinfo{person}{Aaron Baughman}, \bibinfo{person}{Eduardo Morales}, \bibinfo{person}{Rahul Agarwal}, \bibinfo{person}{Gozde Akay}, \bibinfo{person}{Rogerio Feris}, \bibinfo{person}{Tony Johnson}, \bibinfo{person}{Stephen Hammer}, {and} \bibinfo{person}{Leonid Karlinsky}.} \bibinfo{year}{2024}\natexlab{}.
\newblock \showarticletitle{Large scale generative AI text applied to sports and music}. In \bibinfo{booktitle}{\emph{Proceedings of the 30th ACM SIGKDD Conference on Knowledge Discovery and Data Mining}}. \bibinfo{pages}{4784--4792}.
\newblock


\bibitem[Dutta et~al\mbox{.}(2024)]%
        {dutta2024enhancing}
\bibfield{author}{\bibinfo{person}{Arnab Dutta}, \bibinfo{person}{Gleb Polushin}, \bibinfo{person}{Xiaoshuang Zhang}, {and} \bibinfo{person}{Daniel Stein}.} \bibinfo{year}{2024}\natexlab{}.
\newblock \showarticletitle{Enhancing E-commerce Spelling Correction with Fine-Tuned Transformer Models}. In \bibinfo{booktitle}{\emph{Proceedings of the 30th ACM SIGKDD Conference on Knowledge Discovery and Data Mining}}. \bibinfo{pages}{4928--4938}.
\newblock


\bibitem[Fischer et~al\mbox{.}(2024)]%
        {fischer2024grillbot}
\bibfield{author}{\bibinfo{person}{Sophie Fischer}, \bibinfo{person}{Carlos Gemmell}, \bibinfo{person}{Niklas Tecklenburg}, \bibinfo{person}{Iain Mackie}, \bibinfo{person}{Federico Rossetto}, {and} \bibinfo{person}{Jeffrey Dalton}.} \bibinfo{year}{2024}\natexlab{}.
\newblock \showarticletitle{GRILLBot in practice: Lessons and tradeoffs deploying large language models for adaptable conversational task assistants}. In \bibinfo{booktitle}{\emph{Proceedings of the 30th ACM SIGKDD Conference on Knowledge Discovery and Data Mining}}. \bibinfo{pages}{4951--4961}.
\newblock


\bibitem[Gholamalinezhad and Khosravi(2020)]%
        {gholamalinezhad2020pooling}
\bibfield{author}{\bibinfo{person}{Hossein Gholamalinezhad} {and} \bibinfo{person}{Hossein Khosravi}.} \bibinfo{year}{2020}\natexlab{}.
\newblock \showarticletitle{Pooling methods in deep neural networks, a review}.
\newblock \bibinfo{journal}{\emph{arXiv preprint arXiv:2009.07485}} (\bibinfo{year}{2020}).
\newblock


\bibitem[Gholamian et~al\mbox{.}(2024)]%
        {gholamian2024llm}
\bibfield{author}{\bibinfo{person}{Sina Gholamian}, \bibinfo{person}{Gianfranco Romani}, \bibinfo{person}{Bartosz Rudnikowicz}, {and} \bibinfo{person}{Stavroula Skylaki}.} \bibinfo{year}{2024}\natexlab{}.
\newblock \showarticletitle{LLM-Based Robust Product Classification in Commerce and Compliance}. In \bibinfo{booktitle}{\emph{Proceedings of the 1st Workshop on Customizable NLP: Progress and Challenges in Customizing NLP for a Domain, Application, Group, or Individual (CustomNLP4U)}}. \bibinfo{pages}{26--36}.
\newblock


\bibitem[Guo et~al\mbox{.}(2025)]%
        {guo2025deepseek}
\bibfield{author}{\bibinfo{person}{Daya Guo}, \bibinfo{person}{Dejian Yang}, \bibinfo{person}{Haowei Zhang}, \bibinfo{person}{Junxiao Song}, \bibinfo{person}{Ruoyu Zhang}, \bibinfo{person}{Runxin Xu}, \bibinfo{person}{Qihao Zhu}, \bibinfo{person}{Shirong Ma}, \bibinfo{person}{Peiyi Wang}, \bibinfo{person}{Xiao Bi}, {et~al\mbox{.}}} \bibinfo{year}{2025}\natexlab{}.
\newblock \showarticletitle{Deepseek-r1: Incentivizing reasoning capability in llms via reinforcement learning}.
\newblock \bibinfo{journal}{\emph{arXiv preprint arXiv:2501.12948}} (\bibinfo{year}{2025}).
\newblock


\bibitem[Ilharco et~al\mbox{.}(2021)]%
        {ilharco_gabriel_2021_5143773}
\bibfield{author}{\bibinfo{person}{Gabriel Ilharco}, \bibinfo{person}{Mitchell Wortsman}, \bibinfo{person}{Ross Wightman}, \bibinfo{person}{Cade Gordon}, \bibinfo{person}{Nicholas Carlini}, \bibinfo{person}{Rohan Taori}, \bibinfo{person}{Achal Dave}, \bibinfo{person}{Vaishaal Shankar}, \bibinfo{person}{Hongseok Namkoong}, \bibinfo{person}{John Miller}, \bibinfo{person}{Hannaneh Hajishirzi}, \bibinfo{person}{Ali Farhadi}, {and} \bibinfo{person}{Ludwig Schmidt}.} \bibinfo{year}{2021}\natexlab{}.
\newblock \bibinfo{booktitle}{\emph{OpenCLIP}}.
\newblock
\urldef\tempurl%
\url{https://doi.org/10.5281/zenodo.5143773}
\showDOI{\tempurl}
\newblock
\shownote{If you use this software, please cite it as below.}.


\bibitem[Li et~al\mbox{.}(2023)]%
        {li2023labelsupervisedllamafinetuning}
\bibfield{author}{\bibinfo{person}{Zongxi Li}, \bibinfo{person}{Xianming Li}, \bibinfo{person}{Yuzhang Liu}, \bibinfo{person}{Haoran Xie}, \bibinfo{person}{Jing Li}, \bibinfo{person}{Fu lee Wang}, \bibinfo{person}{Qing Li}, {and} \bibinfo{person}{Xiaoqin Zhong}.} \bibinfo{year}{2023}\natexlab{}.
\newblock \bibinfo{title}{Label Supervised LLaMA Finetuning}.
\newblock
\newblock
\showeprint[arxiv]{2310.01208}~[cs.CL]
\urldef\tempurl%
\url{https://arxiv.org/abs/2310.01208}
\showURL{%
\tempurl}


\bibitem[Lokanan and Sharma(2022)]%
        {lokanan2022fraud}
\bibfield{author}{\bibinfo{person}{Mark~Eshwar Lokanan} {and} \bibinfo{person}{Kush Sharma}.} \bibinfo{year}{2022}\natexlab{}.
\newblock \showarticletitle{Fraud prediction using machine learning: The case of investment advisors in Canada}.
\newblock \bibinfo{journal}{\emph{Machine Learning with Applications}}  \bibinfo{volume}{8} (\bibinfo{year}{2022}), \bibinfo{pages}{100269}.
\newblock


\bibitem[Pamina et~al\mbox{.}(2019)]%
        {pamina2019effective}
\bibfield{author}{\bibinfo{person}{Jeyakumar Pamina}, \bibinfo{person}{Beschi Raja}, \bibinfo{person}{S SathyaBama}, \bibinfo{person}{MS Sruthi}, \bibinfo{person}{Aiswaryadevi VJ}, {et~al\mbox{.}}} \bibinfo{year}{2019}\natexlab{}.
\newblock \showarticletitle{An effective classifier for predicting churn in telecommunication}.
\newblock \bibinfo{journal}{\emph{Jour of Adv Research in Dynamical \& Control Systems}}  \bibinfo{volume}{11} (\bibinfo{year}{2019}).
\newblock


\bibitem[Radford et~al\mbox{.}(2021)]%
        {radford2021learning}
\bibfield{author}{\bibinfo{person}{Alec Radford}, \bibinfo{person}{Jong~Wook Kim}, \bibinfo{person}{Chris Hallacy}, \bibinfo{person}{Aditya Ramesh}, \bibinfo{person}{Gabriel Goh}, \bibinfo{person}{Sandhini Agarwal}, \bibinfo{person}{Girish Sastry}, \bibinfo{person}{Amanda Askell}, \bibinfo{person}{Pamela Mishkin}, \bibinfo{person}{Jack Clark}, {et~al\mbox{.}}} \bibinfo{year}{2021}\natexlab{}.
\newblock \showarticletitle{Learning transferable visual models from natural language supervision}. In \bibinfo{booktitle}{\emph{International conference on machine learning}}. PMLR, \bibinfo{pages}{8748--8763}.
\newblock


\bibitem[Roy et~al\mbox{.}(2020)]%
        {roy2020deep}
\bibfield{author}{\bibinfo{person}{Pradeep~Kumar Roy}, \bibinfo{person}{Jyoti~Prakash Singh}, {and} \bibinfo{person}{Snehasish Banerjee}.} \bibinfo{year}{2020}\natexlab{}.
\newblock \showarticletitle{Deep learning to filter SMS Spam}.
\newblock \bibinfo{journal}{\emph{Future Generation Computer Systems}}  \bibinfo{volume}{102} (\bibinfo{year}{2020}), \bibinfo{pages}{524--533}.
\newblock


\bibitem[Sheng et~al\mbox{.}(2024)]%
        {sheng2024measuring}
\bibfield{author}{\bibinfo{person}{Ying Sheng}, \bibinfo{person}{Sudeep Gandhe}, \bibinfo{person}{Bhargav Kanagal}, \bibinfo{person}{Nick Edmonds}, \bibinfo{person}{Zachary Fisher}, \bibinfo{person}{Sandeep Tata}, {and} \bibinfo{person}{Aarush Selvan}.} \bibinfo{year}{2024}\natexlab{}.
\newblock \showarticletitle{Measuring an LLM's Proficiency at using APIs: A Query Generation Strategy}. In \bibinfo{booktitle}{\emph{Proceedings of the 30th ACM SIGKDD Conference on Knowledge Discovery and Data Mining}}. \bibinfo{pages}{5680--5689}.
\newblock


\bibitem[Touvron et~al\mbox{.}(2023)]%
        {touvron2023llama}
\bibfield{author}{\bibinfo{person}{Hugo Touvron}, \bibinfo{person}{Thibaut Lavril}, \bibinfo{person}{Gautier Izacard}, \bibinfo{person}{Xavier Martinet}, \bibinfo{person}{Marie-Anne Lachaux}, \bibinfo{person}{Timoth{\'e}e Lacroix}, \bibinfo{person}{Baptiste Rozi{\`e}re}, \bibinfo{person}{Naman Goyal}, \bibinfo{person}{Eric Hambro}, \bibinfo{person}{Faisal Azhar}, {et~al\mbox{.}}} \bibinfo{year}{2023}\natexlab{}.
\newblock \showarticletitle{Llama: Open and efficient foundation language models}.
\newblock \bibinfo{journal}{\emph{arXiv preprint arXiv:2302.13971}} (\bibinfo{year}{2023}).
\newblock


\bibitem[Wan et~al\mbox{.}(2024)]%
        {wan2024tnt}
\bibfield{author}{\bibinfo{person}{Mengting Wan}, \bibinfo{person}{Tara Safavi}, \bibinfo{person}{Sujay~Kumar Jauhar}, \bibinfo{person}{Yujin Kim}, \bibinfo{person}{Scott Counts}, \bibinfo{person}{Jennifer Neville}, \bibinfo{person}{Siddharth Suri}, \bibinfo{person}{Chirag Shah}, \bibinfo{person}{Ryen~W White}, \bibinfo{person}{Longqi Yang}, {et~al\mbox{.}}} \bibinfo{year}{2024}\natexlab{}.
\newblock \showarticletitle{Tnt-llm: Text mining at scale with large language models}. In \bibinfo{booktitle}{\emph{Proceedings of the 30th ACM SIGKDD Conference on Knowledge Discovery and Data Mining}}. \bibinfo{pages}{5836--5847}.
\newblock


\bibitem[Wang(2023)]%
        {Wang2023CalibrationID}
\bibfield{author}{\bibinfo{person}{Cheng Wang}.} \bibinfo{year}{2023}\natexlab{}.
\newblock \showarticletitle{Calibration in Deep Learning: A Survey of the State-of-the-Art}.
\newblock \bibinfo{journal}{\emph{ArXiv}}  \bibinfo{volume}{abs/2308.01222} (\bibinfo{year}{2023}).
\newblock
\urldef\tempurl%
\url{https://api.semanticscholar.org/CorpusID:260379149}
\showURL{%
\tempurl}


\bibitem[Wang et~al\mbox{.}(2023)]%
        {wang2023prompt}
\bibfield{author}{\bibinfo{person}{Yuki Wang}, \bibinfo{person}{Wei Wang}, \bibinfo{person}{Qi Chen}, \bibinfo{person}{Kaizhu Huang}, \bibinfo{person}{Anh Nguyen}, {and} \bibinfo{person}{Suparna De}.} \bibinfo{year}{2023}\natexlab{}.
\newblock \showarticletitle{Prompt-based zero-shot text classification with conceptual knowledge}. In \bibinfo{booktitle}{\emph{Proceedings of the 61st Annual Meeting of the Association for Computational Linguistics (Volume 4: Student Research Workshop)}}, Vol.~\bibinfo{volume}{4}. Association for Computational Linguistics, \bibinfo{pages}{30--38}.
\newblock


\bibitem[Xu et~al\mbox{.}(2024)]%
        {xu2024llms}
\bibfield{author}{\bibinfo{person}{Hanzi Xu}, \bibinfo{person}{Renze Lou}, \bibinfo{person}{Jiangshu Du}, \bibinfo{person}{Vahid Mahzoon}, \bibinfo{person}{Elmira Talebianaraki}, \bibinfo{person}{Zhuoan Zhou}, \bibinfo{person}{Elizabeth Garrison}, \bibinfo{person}{Slobodan Vucetic}, {and} \bibinfo{person}{Wenpeng Yin}.} \bibinfo{year}{2024}\natexlab{}.
\newblock \showarticletitle{LLMs' Classification Performance is Overclaimed}.
\newblock \bibinfo{journal}{\emph{arXiv preprint arXiv:2406.16203}} (\bibinfo{year}{2024}).
\newblock


\bibitem[Zhu et~al\mbox{.}(2023)]%
        {zhu2023prompt}
\bibfield{author}{\bibinfo{person}{Yi Zhu}, \bibinfo{person}{Ye Wang}, \bibinfo{person}{Jipeng Qiang}, {and} \bibinfo{person}{Xindong Wu}.} \bibinfo{year}{2023}\natexlab{}.
\newblock \showarticletitle{Prompt-learning for short text classification}.
\newblock \bibinfo{journal}{\emph{IEEE Transactions on Knowledge and Data Engineering}} (\bibinfo{year}{2023}).
\newblock


\end{thebibliography}
\appendix  

\section{Comparison of GPT-4o with GPT-4o mini}
\label{app:openai_comparison}
\begin{figure}[h]
\begin{center}
\includegraphics[width=0.48\textwidth]{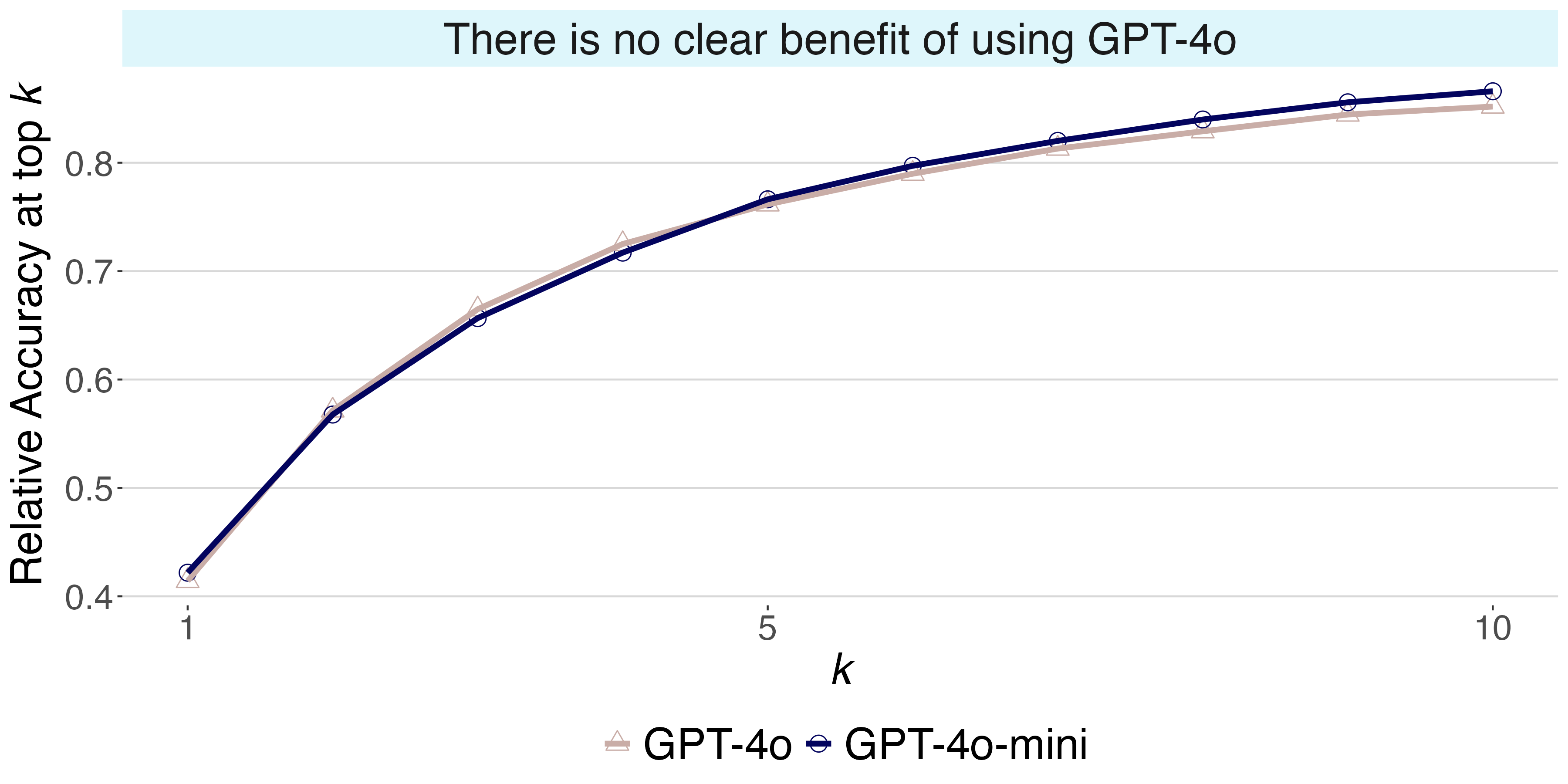}
\vspace{-0.3cm}
\caption{Comparison of OpenAI models shows that there is no clear benefit in using GPT-4o over GPT-4o mini for our task.}
\label{fig:openai_mini}
\end{center}
\end{figure}

Figure~\ref{fig:openai_mini} shows that for our use case, there is no difference in performance between OpenAI's GPT 4o and GPT 4o mini.

\section{Accuracy comparison across different types of inputs}
\label{app:inputs}

\begin{figure*}[h]
\begin{center}
\includegraphics[width=0.96\textwidth]{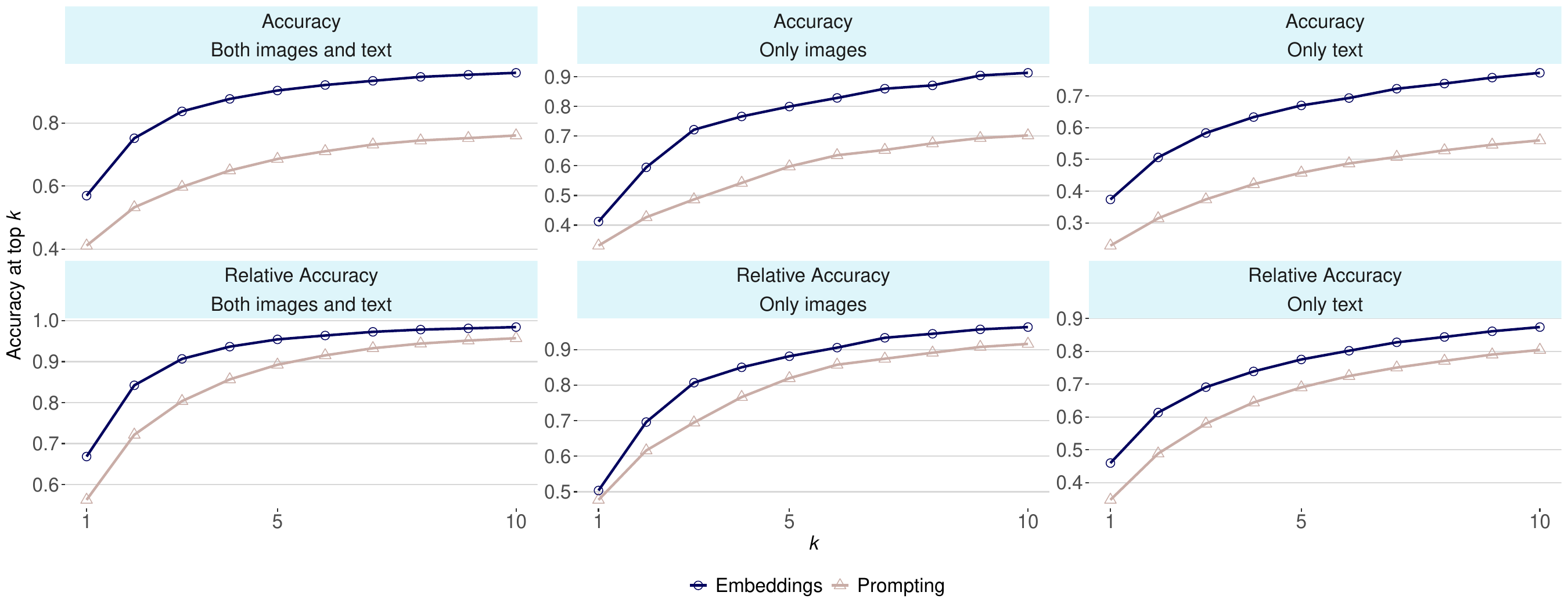}
\vspace{-0.3cm}
\caption{Accuracy for different input types: across all tests and both accuracy metrics (Equations~\ref{eq:acc} and \ref{eq:rel}) the embeddings approach performs significantly better than prompting.}
\label{fig:text_image}
\end{center}
\end{figure*}

Figure~\ref{fig:text_image} in the next page breaks down the performance of the two approaches across different inputs. The top row shows the accuracy (Equation~\ref{eq:acc}) while the bottom row shows the relative accuracy (Equation~\ref{eq:rel}). As we discussed in Section~\ref{sec:results}, across all categories the embeddings approach significantly outperforms prompting.

\end{document}